\documentclass{bmvc2k}
\usepackage{booktabs} 
\usepackage{multirow}
\usepackage{amssymb}
\usepackage{capt-of}

\title{HR-RCNN: Hierarchical Relational Reasoning for Object Detection}

\addauthor{Hao Chen, Abhinav Shrivastava}{{chenh, abhinav}@cs.umd.edu}{1}

\addinstitution{
 University of Maryland\\
 College Park, USA
}

\runninghead{Chen \& Shrivastava}{HR-RCNN}

\def\eg{\emph{e.g}\bmvaOneDot}

\def\etal{\emph{et al}\bmvaOneDot}
\def\ie{\emph{i.e}\bmvaOneDot} 
 
\def\etc{\emph{etc}\bmvaOneDot} \def\vs{\emph{vs}\bmvaOneDot~}
 
\def\etal{\emph{et al}\bmvaOneDot}

\begin{document}

\maketitle

\begin{abstract}
Incorporating relational reasoning in neural networks for object recognition remains an open problem. Although many attempts have been made for relational reasoning, they generally only consider a single type of relationship. For example, pixel relations through self-attention (e.g., non-local networks), scale relations through feature fusion (e.g., feature pyramid networks), or object relations through graph convolutions (e.g., reasoning-RCNN). Little attention has been given to more generalized frameworks that can reason across these relationships. In this paper, we propose a hierarchical relational reasoning framework (HR-RCNN) for object detection, which utilizes a novel graph attention module (GAM). This GAM is a concise module that enables reasoning across heterogeneous nodes by operating on the graph’s edges directly. Leveraging heterogeneous relationships, our HR-RCNN shows great improvement on COCO dataset, for both object detection and instance segmentation.

\end{abstract}

\section{Introduction}

Even though convolutional neural networks (ConvNets)~\cite{alexnet,vgg,resnet}  have revolutionized the field of object recognition in recent years~\cite{ren2015faster,lin2017feature,he2017mask,lin2017focal,shrivastava2016beyond,ohem_cvpr16}, they are still fairly limited in their ability to \emph{explicitly} model and reason about a myriad of contextual relationships in images~\cite{biederman1981semantics}. 
Standard feedforward ConvNets (\cite{alexnet,vgg}) rely on a data-driven approach to implicitly learn these relationships, \ie, given enough data, the model should learn whatever is necessary implicitly without explicitly modeling context. This is contrary to human visual pathways, which rely heavily on context and relational reasoning for perception~\cite{biederman1981semantics,palmer1975effects,hock1974contextual,torralba2003context}. 
To address this, many efforts have been made to encode context and relational information in ConvNets, ranging from designing architectures that better learn relationships to explicitly modeling relations by utilizing graph formulations.

\begin{figure}[t]
    \centering
    \includegraphics[width=.65\linewidth]{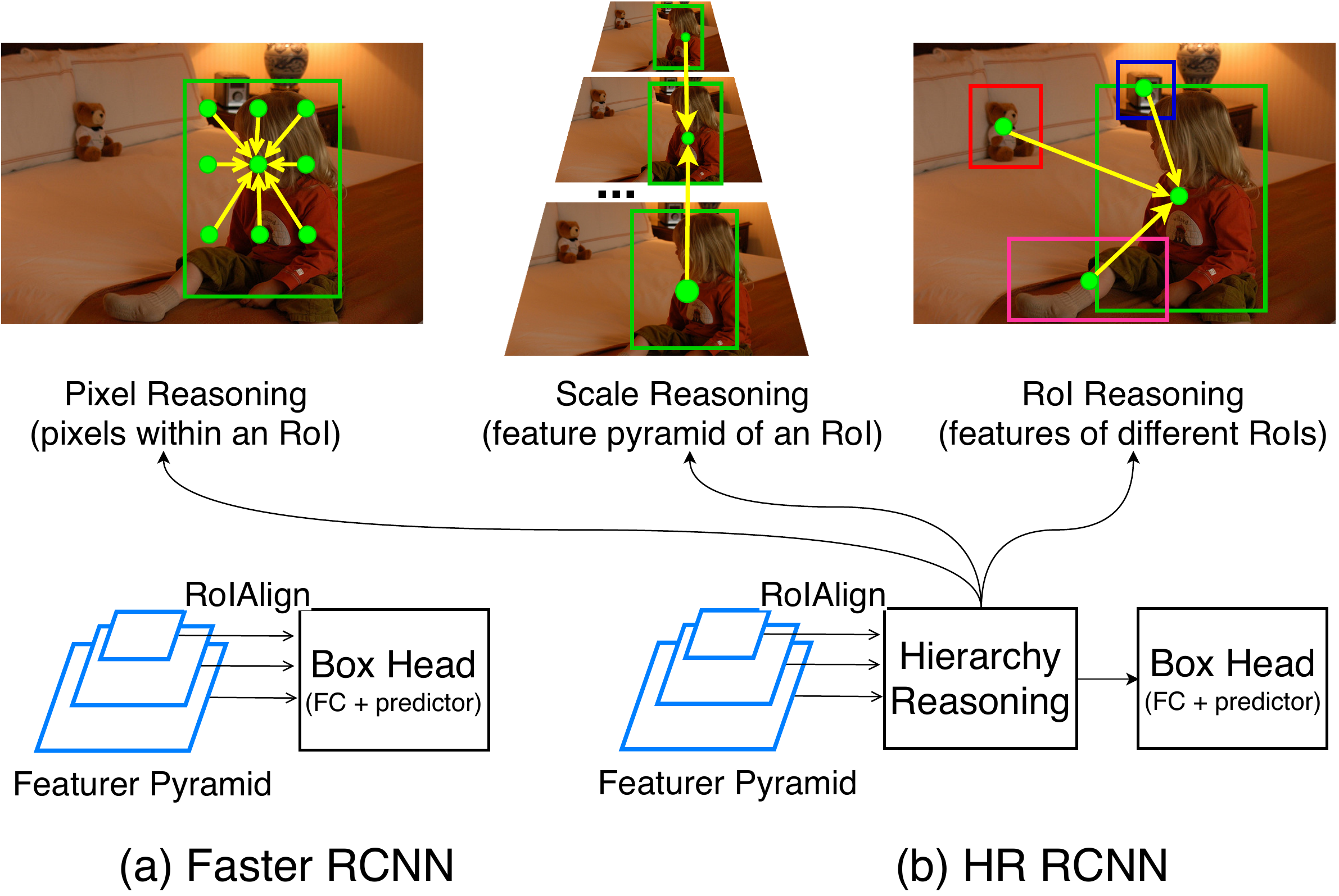}
    \centering
    \caption{The Hierarchical Relational Reasoning Framework for object detection, where pixel relationships, scale relationships, RoI relationships are incorporated in one network.}
    \label{fig:idea}
    \vspace{-0.1cm}
\end{figure}

We  characterize the different relation information into three groups: pixel relations, scale relations, and object relations. For modeling pixel relations, conditional random fields (CRF)~\cite{krahenbuhl2011efficient} model interactions between all (or selected) pairs of pixels in an image and have been successfully integrated into neural networks widely~\cite{zheng2015conditional,arnab2016higher,arnab2018conditional,xu2017learning}. Similarly, works on non-local relationships~\cite{wang2018nonlocal,cao2019gcnet,huang2018ccnet,Zhang2020DynamicGM} utilize a self-attention mechanism to model relationships. Both these approaches rely on assimilating information via their pixel-connectivity to improve feature representations.
For scale relations, many efforts have been made on fusing features across scales to alleviate the discrepancy of feature maps from different levels of bottom-up hierarchy and feature scale-space, including top-down information flow~\cite{shrivastava2016beyond,Lin_2017_CVPR,fu2017dssd}, an extra bottom-up information  path~\cite{liu2018path,woo2019gated,kong2017ron}, multiple hourglass structures~\cite{newell2016stacked,zhao2019m2det}, concatenating features from different layers~\cite{li2017fssd,sun2019high,hariharan2015hypercolumns,bell2016inside} or different tasks~\cite{primingfeedback_eccv16}, gradual multi-stage local information fusions ~\cite{yu2018deep,sun2019deep}, pyramid convolutions~\cite{wang2020scale}, \etc. Even though standard design principles for scale relations are emerging for ConvNet architectures, the problem is far from being solved. For object relations, some early works like~\cite{kipf2016semi,marino2016more,jiang2018hybrid,chen2018iterative} used annotated relationships or handcrafted linguistic knowledge to build explicit relationships between object classes; and more recently, several methods~\cite{liu2018structure,chen2017spatial,hu2018relation,xu2019spatial,xu2019reasoning} learn the relational graph between different regions' visual features to improve the handcrafted object relation graph further. Although all these approaches attempt to embed relational reasoning one way or another, little attention has been given to a more generic and unified model that can integrate these seamlessly.
Moreover, these relations are not independent, but their dependencies are not well established. Therefore, we propose an approach to unify these different contextual relations (\emph{viz.}, pixel, scale, and object relations) in a single model, that also provides a principled way to explore relational hierarchies.

In this paper, we present a Hierarchical Relational framework for object detection (HR-RCNN), which is illustrated in Fig.~\ref{fig:idea}.
We build on a Faster R-CNN (Fig.~\ref{fig:idea}(a)) detection model, where a backbone network extracts feature pyramid and generates region proposals for an image, the per-region features are extracted from a specific level of the feature pyramid, and these features are input to a box head and processed separately. In contrast to this paradigm, when given region proposals and a feature pyramid, our HR-RCNN (Fig.~\ref{fig:idea}(b)) inserts a hierarchical relational reasoning (HR) component between the feature pyramid and box head. Specifically, we embed three relational reasoning components in the HR component: a pixel graph, a scale graph, and a region-of-interest (RoI) graph. To simplify and unify the model architecture, we design a novel and concise graph attention module (GAM), illustrated in Fig.~\ref{fig:gar} which can assimilate information from heterogeneous graphs. In GAM, every node encodes the semantic and spatial distance from its neighbors into edges, and outputs attention weights by directly operating on the edges.
Then, following standard graph neural network (GNN) approaches~\cite{hamilton2017inductive, kipf2016semi,hamaguchi2017knowledge,khalil2017learning,sanchez2018graph,battaglia2016interaction}, we enhance the node feature by assimilating messages from its neighbors weighted by their respective relations. By reasoning through heterogeneous nodes, our GAM aggregates information from the whole graph and leads to a refined representation. Since we have multiple relation graphs, we build a hierarchy between these relations. and it leads to our final HR-RCNN architecture. We conduct extensive experiments on the COCO object detection and instance segmentation dataset to demonstrate the efficiency of our approach and provide a comprehensive analysis.

We summarize our contributions into three parts.
\textbf{First}, we propose a hierarchical relational reasoning framework which integrates pixel relations, scale relations, and RoI relations into a single model.
\textbf{Second}, we design a concise graph attention module for visual reasoning and explore different ways of build hierarchies of relation graphs.
\textbf{Finally}, we demonstrate consistent improvements with HR-RCNN on COCO dataset.

\section{Related Works}

\smallskip\noindent\textbf{Pixel relations.}
Encoding context information through pixel relation has a long history in computer vision. The standard paradigm before deep learning was the algorithm by~\cite{krahenbuhl2011efficient} that attempted to model pixel relation of all pixel pairs through conditional random fields~(CRF), under mild assumptions. And while~\cite{zheng2015conditional,arnab2016higher,arnab2018conditional,xu2017learning} utilize similar CRF as a separate module in their neural networks, the process remains cumbersome and computationally expensive. Therefore, learning-based approaches have recently drawn more attention. In that space, deformable convolution~\cite{dai2017deformable, zhu2019deformable} learns the offsets with respect to a predefined grid to generate content-adaptive inputs. Self-attention methods~\cite{wang2018nonlocal,cao2019gcnet,huang2018ccnet,Zhang2020DynamicGM}, on the other hand, model pairwise relationships and generate attention weights through scaled-dot-product. While most pixel relation modules are adopted in the backbone for scene understanding of the whole image, our pixel graph is built only from the pixel features within an RoI.

\smallskip\noindent\textbf{Scale relations.}
The literature to learn scale invariance and relationships across scales can be divided into two groups: image pyramids and feature pyramids. In recent years, image pyramid approaches, like SNIP~\cite{singh2018analysis} and SNIPER~\cite{singh2018sniper}, introduce scale normalization to improve the performance efficiently. 
Approaches that model feature pyramids attempt to fuse information from low-level features (rich in details) and high-level features (rich in semantic information). Towards this, TDM~\cite{shrivastava2016beyond} and Feature Pyramid Networks (FPN)~\cite{lin2017feature} introduce a top-down and lateral connections to integrate information of multiple levels and thus passing on semantic information into low-level features.
Based on FPN, many structure enhancement are proposed by PANet~\cite{liu2018path}, Bi-FPN~\cite{Tan_2020_CVPR}, NAS-FPN~\cite{ghiasi2019fpn}, SEPC~\cite{wang2020scale}, and DetectorRS~\cite{qiao2020detectors}. For HR-RCNN, we incorporate the scale relations in a scale graph, where nodes are different levels of corresponding features in the feature pyramid.

\smallskip\noindent\textbf{Object relations.}
Visual reasoning from the interaction or dependency information between objects has been widely studied across a wide range of vision tasks, such as image classification~\cite{marino2016more}, scene parsing~\cite{zellers2018neural}, scene graph generation~\cite{gu2019scene,li2018factorizable} and large-scale object detection~\cite{chen2018iterative,xu2019reasoning,xu2019spatial}.
Object relation attempts to model contextual relationships between objects for visual understanding. Recent works utilize such relationships by an explicit handcrafted knowledge graph~\cite{kipf2016semi,marino2016more,jiang2018hybrid,chen2018iterative}, or an implicit learning graph~\cite{liu2018structure,chen2017spatial}.
For object detection, the literature has explored region-region relationships, class-region relationships, and class-class relationships. For example, relation network~\cite{hu2018relation} incorporates region-region relationships using self-attention, SGRN~\cite{xu2019spatial} reasons based on a spatial-aware region-region graph including both appearance and spatial dependencies, Reasoning-RCNN~\cite{xu2019reasoning} builds a class-class graph with global image-wise information, and Chen, Xinlei~\etal~\cite{chen2018iterative} iteratively stacks multiple graphs in a reasoning framework.
Compared to these, our HR-RCNN encodes three levels of relationships (pixel, scale, and RoI) in a unified framework.

\smallskip\noindent\textbf{Graph attention network.}
Built upon graph convolution networks (GCN), graph attention networks (GAT)~\cite{velivckovic2017graph} proposes a self-attention framework for any type of structure data. In GAT, every node is assigned an attention weight, which is used in the following feature aggregation. Due to its generalization and effectiveness, GAT has been utilized in many fields, such as point cloud instance segmentation~\cite{wang2019graph}, visual question answering~\cite{li2019relation}, trajectory forecasting~\cite{kosaraju2019social}, and referring expression comprehension~\cite{yang2019dynamic}. Similar to GAT~\cite{velivckovic2017graph}, our graph attention module (GAM) collects information from heterogeneous nodes by a corresponding attention weight. In our GAM, we concatenate the semantic and spatial relationships into the edges and operate on the edge directly to output the attention weight.

\smallskip\noindent\textbf{Other attention detectors.}
RelationNet++\cite{relationnetplusplus2020} combines heterogeneous visual representations~(\eg, representations for anchor box, region proposal, corner/center points) with an efficient bridging visual representations (BVR), via key sampling and shared location embedding.
HoughNet\cite{HoughNet}, as a voting-based bottom-up object detector, integrates local and long-range context information for object localization.
Dynamic Head~\cite{dai2021dyhead} also integrates multiple attention modules, but their attention implementation are quite different for three different parts: linear function for scale-aware attention, deformable convolution for spatial location sampling, and a gating subnetwork for task-aware attention in channel dimension. Also, They stack attention blocks multiple times to boost the feature representation. We provide comparison with DyHead in the supplementary material.

\section{Our Approach}

\noindent\textbf{Preliminaries.} 
Firstly, we briefly revisit the architecture of a region-based object detector (illustrated in Fig.~\ref{fig:idea}(a)).
The detection network can be divided into three parts: a backbone network for feature extraction, a region proposal network (RPN) for proposal generation, and a box head for final classification and localization.
To enable hierarchical reasoning, we plug three visual reasoning modules between the backbone and box head, using a concise graph attention module (GAM). In GAM, all relationships are embedded in the edge attribute, and we operate directly on the edges to output the attention weights for feature enhancement. Finally, we combine heterogeneous visual reasoning modules into our HR-RCNN.

\medskip\noindent\textbf{Graph construction}
Firstly, we construct graphs for visual reasoning. For pixel graph, we use a single-pixel within an RoI as a node, and a fully-connected graph is the feature representation for the RoI. For scale graph, we use an RoI-pooled~\cite{ren2015faster} feature from a particular feature pyramid level as a node, and the graph representing the RoI connects these feature nodes across different levels. Finally, for the RoI graph, we use an RoI's feature as a node, and the graph is constructed by connecting nodes from different RoIs in an image. To fully leverage the relationships, all graphs are fully-connected and adaptively assign attention weights to their neighbors.

\begin{figure}[t!]
    \centering
      \includegraphics[width=.98\linewidth]{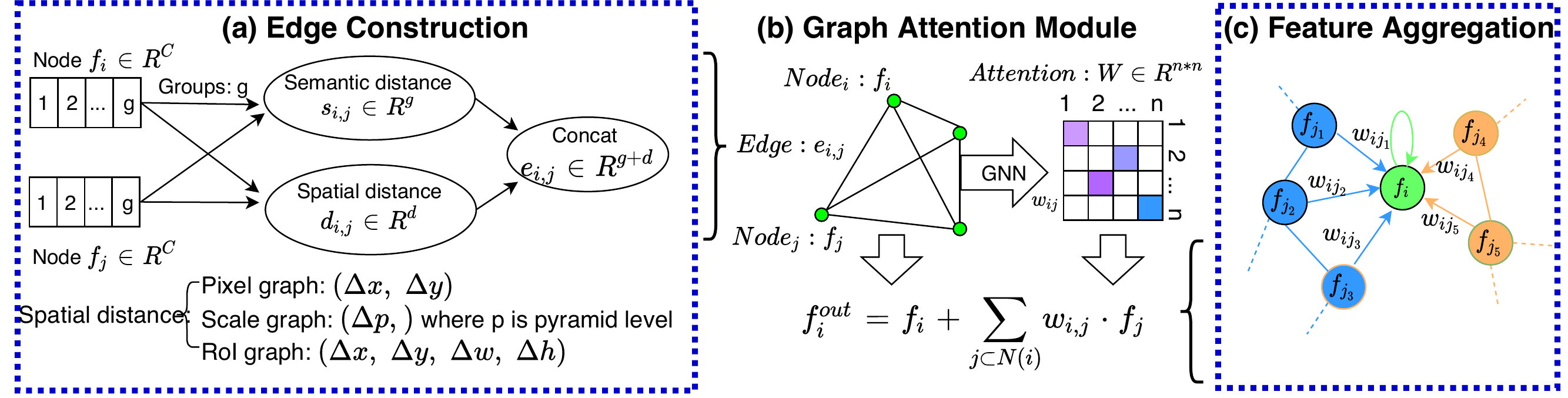}
    \caption{Graph attention module (GAM).}
    \label{fig:gar}
\end{figure}

Edge attribute, which models the relationship between nodes, can be divided into two parts: semantic distance measuring node distance in the feature space and spatial distance measuring node distance in the spatial space ( Fig.~\ref{fig:gar}(a)). For semantic distance $s_{i,j}$, following \cite{wang2018nonlocal,cao2019gcnet,huang2018ccnet}, we first divide the node attribute into $g$ groups and then compute their semantic distance by groups,
$s_{ij} = \text{SemanticDistance}(f_i, f_j),$
where $i$ is the query node and $j$ is the key node.
The semantic distance can be implemented in many ways, including dot product, cosine similarity, or euclidean distance.

For spatial distance $d_{i,j}$, different relation graphs have their own definitions. Pixel graph uses normalized  $(x, y)$ distance in the pixel space:
\begin{math}
d^{P}_{i,j} = (\Delta x, \Delta y) = \left(\frac{x_i - x_j}{w}, \frac{y_i - y_j}{h}\right),
\end{math}
where $w,h$ is the RoI width and height.
Scale graph uses the normalized level distance in the feature pyramid space:
\begin{math}
d^{S}_{i,j} = (\Delta p) = \left(\frac{p_i - p_j}{P}\right),
\end{math}
where $P$ is the number of pyramid levels. The RoI graph uses normalized $(x_\text{center},y_\text{center})$, width, and height distance:
\begin{math}
d^{R}_{i,j} = (\Delta x, \Delta y, \Delta w, \Delta h ) = \left(\frac{x_i - x_j}{w_i}, \frac{y_i - y_j}{h_i}, \frac{w_j}{w_i}, \frac{h_j}{h_i}\right).
\end{math}
Finally, we concatenate the semantic distance and spatial distance as the edge attribute. An illustration of edge construction is shown in Fig~\ref{fig:gar} (a).

\medskip\noindent\textbf{Graph Attention Module}
Once we have the relation graph constructed, a message-passing strategy needs to be designed for relational reasoning. Similar to graph attention network (GAT)~\cite{velivckovic2017graph}, our Graph Attention Module (GAM, illustrated in Fig~\ref{fig:gar}(b)) generates attention weights for all neighbor nodes, and a weighted sum operation is utilized for feature aggregation. Since we already collect all the relationships in the edge attributes, our GAM operates directly on edges to produce the attention weights. Specifically, we introduce a  multi-layer perceptron (\texttt{mlp}) to generate the attention weight $\alpha_{ij}$: $\alpha_{ij} = \text{\texttt{mlp}}(e_{ij}),$
which is followed by a global reasoning strategy, where we normalize the attention weights for all its neighbors.
For simplicity, we implement this step as a softmax function, 
\begin{equation}
        w_{ij} = \text{softmax}_j\left(\alpha_{ij}/T\right) = \frac{\exp\left(\alpha_{ij}/T\right)}{\sum\limits_{k \in N(i)} \exp\left(\alpha_{ik}/T\right)},
    \label{equ:softmax}
\end{equation}

where $w_{ij}$ is the normalized attention weight, $T$ is the softmax temperature (set as $2$), $N(i)$ is the neighbor nodes for query node $i$.
Finally, we refine the node features by aggregating information from the graph structure and all neighbors. Specifically, we enhance the node feature using residual connections, where the enhancement comes from the weighted sum of neighbor nodes (illustrated in Fig~\ref{fig:gar}(c)).
\begin{equation}
    f_i^\text{out} = f_i + \sum\limits_{j\in N(i)} w_{ij}  f_j.
\end{equation}
After we obtain the updated node features, an additional FC layer is added for further fusion.

\begin{figure}[t!]
    \centering
    \includegraphics[width=.65\linewidth]{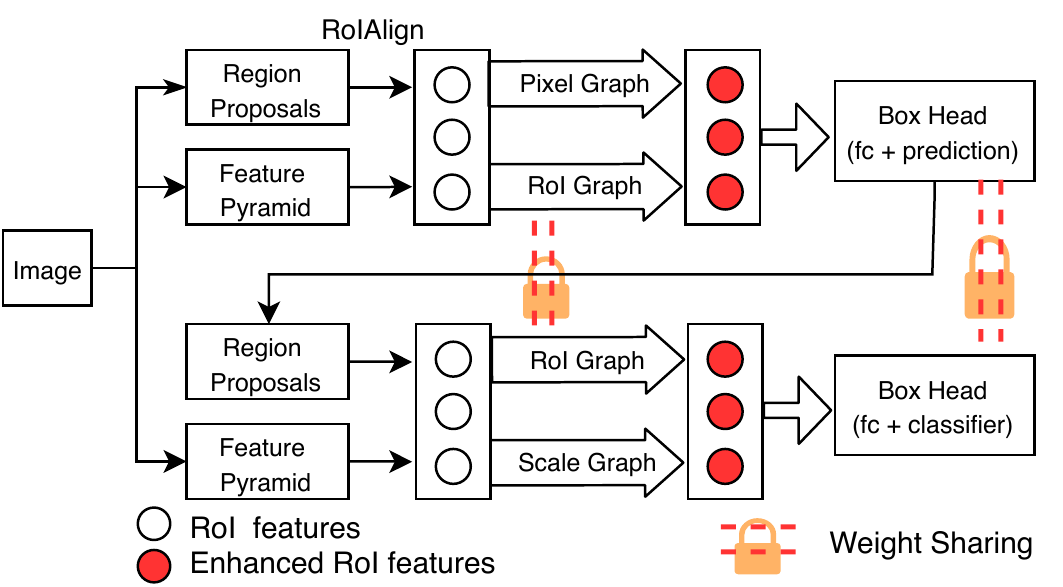}
    \caption{The proposed HR-RCNN structure.}
    \label{fig:refine}  
\end{figure}

\medskip
\noindent\textbf{Hierarchical Relation Reasoning}
We propose our hierarchical reasoning framework based on a hierarchy of different relationships~(Figure~\ref{fig:refine}). Given the context it captures, we build two reasoning modules using these three graphs: an intra-level and inter-level context reasoning module, where level is a feature pyramid level. Both these modules use the RoI graph jointly with pixel and scale graphs respectively. The first stage embeds pixel graph and RoI graph jointly, while the second stage embeds scale graph and RoI graph jointly. Finally, to better utilize the learned knowledge from the first reasoning stage, we also utilize first stage's predictions as the region proposals for the second stage, \ie, iterative bounding-box regression~\cite{ohem_cvpr16,bell2016inside}). This leads to our HR-RCNN framework (illustrated in Fig.~\ref{fig:refine}), where hierarchical reasoning happens in two stages. We train all reasoning branches jointly. This particular design is partly inspired by empirical findings on the complementary nature of different relation graphs, presented in Tab.~\ref{tab:further-ablation1} and supplementary material.

\section{Experiment}
\smallskip
\noindent\textbf{Datasets and implementation details.}
We conduct all experiments on the  COCO 2017 dataset~\cite{lin2014microsoft}, with train split ($\sim$118k images) for training, val~($\sim$5k images) split and test-dev split~($\sim$ 20k images, annotations withheld) for evaluation.
All experiments are implemented using Detectron2~\cite{wu2019detectron2}. The input images are resized to have a shorter size of 800 pixels while the longer side no more than 1333 pixels. By default, we train the models with a total of 16 images per minibatch on 4 GPUs. Unless otherwise specified,  all models are trained for 90k iterations (denoted as $1\times$ lr\_scheduler) with an initial learning rate of 0.02, decreasing by a ratio of 0.1 at 60k and 80k respectively. We utilize ResNet-50 and ResNet-101 with feature pyramid network as backbones, and the batch normalization layers are fixed  during training. All other hyper-parameters in this paper follow the settings in Detectron2. For HR-RCNN, we set group size for semantic distance to 2. 
\textit{More results, graph combinations, other backbone main results, temperature/group size ablations, attention weights and detection results visualisation \etc, can be found in the supplementary material}.

\subsection{Main Results}
\begin{table}[t!]
\centering
\footnotesize
\caption{\textbf{Main Results} on COCO validation set. The impact of using HR-RCNN with different backbones. All methods are based on Faster RCNN with feature pyramid network.}
\label{tab:setups}
\begin{tabular}{@{}lcccccc@{}}
\toprule
 Methods             & AP   & AP$_{50}$ & AP$_{75}$ & AP$_\text{S}$  & AP$_\text{M}$  & AP$_\text{L}$  \\
\midrule                
ResNet50~\cite{resnet}           & 38.0   & 58.6 & 41.4 & 22.1 & 41.8 & 48.8 \\
R50\_cascade      & 40.4 & 60.2 & 43.8 & 23.8 & 44.1 & 52.2 \\
HR-RCNN   & \textbf{41.6} & \textbf{61.8} & \textbf{45.2} & \textbf{25}   & \textbf{45.1} & \textbf{54.2} \\ 
\midrule                
ResNet101~\cite{resnet}       & 40.2 & 61.2 & 43.8 & 24.1 & 43.8 & 52.1 \\
R101\_cascade     & 42.1 & 62.3 & 45.4 & 24.7 & 45.5 & 54.2 \\
HR-RCNN   & \textbf{42.8} & \textbf{63.1} & \textbf{46.3} & \textbf{25.5} & \textbf{46.4} & \textbf{55.8} \\ 
\midrule                
DCN-V2~\cite{zhu2019deformable} & 40.8 & 62.0   & 44.5 & 24.2 & 44.0   & 54.0   \\
R50\_DCN\_cascade & 42.3 & 62.5 & 45.7 & 25.8 & 45.1 & 56.2 \\
HR-RCNN   & \textbf{42.9} & \textbf{63.2} & \textbf{46.6} & \textbf{26.2} & \textbf{45.9} & \textbf{57.1} \\ 
\bottomrule
\end{tabular}
\end{table}

\begin{table}[]
\centering
\caption{\textbf{Hierarchical  reasoning \vs single-level reasoning}. Methods with * means using segmentation annotations,  with $\dagger$ means trained for $2\times$ epochs.}
\label{tab:relation}
\footnotesize
\begin{tabular}{@{}l|l|ccc|ccc@{}}
\toprule
   Methods                     &Backbone  & Pixel & Scale & RoI & AP   & AP$_{50}$ & AP$_{75}$\\
\midrule                      
GCNet~\cite{cao2019gcnet}*               &Res50 & \checkmark       &       &     & 38.7    & 61.1 & 41.7 \\
Non-local~\cite{wang2018nonlocal}*           &Res50 &\checkmark        &       &     & 39.0      & 61.1 & 41.9 \\
DCN-V2~\cite{zhu2019deformable}*           &Res50 &\checkmark        &       &     & 39.9      & - & - \\
DGMN~\cite{Zhang2020DynamicGM}*                &Res50 &\checkmark        &       &     & 40.2    & 62.0   & 43.4 \\
AugFPN~\cite{fan2020augfpn}				&Res50 &       &   \checkmark     &  &38.8	 &61.5	&42.0 \\
SEPC~\cite{wang2020scale}                 &Res50 &       &   \checkmark     &     & 38.5    & 59.9 & 41.4 \\
NAS-FPN~\cite{ghiasi2019fpn}              &Res50 &       & \checkmark        &     & 39.7    & 57   & 41.8 \\
RelationNet~\cite{hu2018relation}          &Res50 &       &       &    \checkmark  & 38.8    & 60.3 & 42.9 \\
HR-RCNN (Ours)        &Res50 & \checkmark       &  \checkmark      & \checkmark     & \textbf{41.6}    & \textbf{61.8} & \textbf{45.2} \\
\midrule
SGRN~\cite{xu2019spatial} $\dagger$         &Res101   &       &       &  \checkmark    & 41.7    & 62.3 & 45.5 \\
Reasoning-RCNN~\cite{xu2019reasoning} $\dagger$ &Res101 &       &       & \checkmark     & 42.9    & -    & -    \\
HR-RCNN$\dagger$  (Ours) &Res101   &   \checkmark     & \checkmark       & \checkmark     & \textbf{44.6}    & \textbf{64.7} & \textbf{ 48.1} \\
\bottomrule
\end{tabular}
\end{table}

\noindent\textbf{Generalization across backbones.}
In this section, we evaluate HR-RCNN on COCO val2017 set with backbone different architectures, including FPN~\cite{lin2017feature} with ResNet-50~\cite{resnet}, ResNet-101~\cite{resnet}, Deformable ResNet-50~\cite{zhu2019deformable}.
For fair comparisons, we report our re-implemented results as the baseline; \ie, our Faster R-CNN implementation with these backbones (which are generally better than originally reported) and then HR-CNN applied on that model.
Since HR-RCNN has one more stage than the baseline, we also include cascade RCNN here.
These results are shown in Tab.~\ref{tab:setups}, where HR-RCNN consistently improves the performance for ResNet in all metrics. 
Note that all our backbones utilize the FPN structure, which already considers scale relation to address the scale invariance. This demonstrates that our approach is complementary to the feature pyramid architectures. 

\smallskip\noindent\textbf{Hierarchical \vs Single-level  Reasoning.}
To show the advantage of hierarchical reasoning over single-level reasoning, we compare with many single-level reasoning methods in Tab.~\ref{tab:relation}.
A clear AP improvement can be seen from hierarchical visual reasoning as compared to single-level reasoning.
Therefore, by iteratively extracting heterogeneous relationships, our HR-RCNN framework can gradually enhance the feature representation of stacked ConvNets and greatly improve the final performance.

\smallskip\noindent\textbf{Instance Segmentation}
To further evaluate the hierarchical reasoning framework, we extend it to the instance segmentation task. We take the Mask RCNN~\cite{he2017mask} with ResNet50-FPN backbone as the instance segmentation baseline. To incorporate hierarchical visual reasoning, we keep the box head same as HR-RCNN and put a single P$+$R reasoning component in the mask head. As shown in Tab.~\ref{tab:seg}, we improve the segmentation AP by 1.9 points, which shows the potential of hierarchical visual reasoning to other tasks.

\begin{table}[t!]
    \centering
    \footnotesize
    \captionof{table}{\textbf{Instance segmentation} results}
    \label{tab:seg}
        \begin{tabular}{@{}lcccccc@{}}
        \toprule
                              & \multicolumn{3}{c}{Box}               & \multicolumn{3}{c}{Segmentation}         \\
        \cmidrule(lr){2-4}
        \cmidrule(lr){5-7}
                      & AP            & AP$_{50}$     & AP$_{75}$     & AP            & AP$_{50}$     & AP$_{75}$     \\
        \midrule              
        Mask RCNN & 38.6          & 59.5          & 42.1          & 35.2          & 56.3          & 37.5          \\
        Cascade Mask RCNN & 41.3         & 60.1          & 45.1          & 36.1         & 57.3          & 38.8          \\
                    
        HR-Mask RCNN  & \textbf{41.8} & \textbf{61.7} & \textbf{45.4} & \textbf{37.1} & \textbf{58.6} & \textbf{39.7} \\ 
        \bottomrule
        \end{tabular}
\end{table}

\subsection{Ablation Study}
\label{sec:ablation}

\begin{table}[t]
    \centering
    \footnotesize
    \caption{Ablations results for \textbf{individual relations}}
    \label{tab:single-relation}
    \begin{tabular}{@{}l|cc|c|cccccc@{}}
    \toprule
                & Semantic & Spatial & $\Delta$ params &AP  & AP$_{50}$ & AP$_{75}$ & AP$_\text{S}$  & AP$_\text{M}$  & AP$_\text{L}$  \\
    \midrule
    Baseline    &   -      &    -     & - & 38.0   & 58.6 & 41.4 & 22.1 & 41.8 & 48.8 \\
    \midrule
    \multirow{2}{*}{Pixel graph}            & \checkmark         &      &64    & 38.3 & 59.1 & 41.6 & 22.2 & 41.6 & 49.8 \\
     &  \checkmark        &   \checkmark     & 608   & 38.5 & 59.3 & 41.9 & 22.5 & 42.1 & 49.9 \\
    \midrule
    \multirow{2}{*}{Scale graph}            & \checkmark         &      &96    & 38.2 & 58.9 & 41.4 & 22.2 & 41.5 & 49.2 \\
     &  \checkmark        &   \checkmark       & 192 & 38.5 & 59.3 & 41.7 & 22.4 & 42   & 49.5 \\
    \midrule
    \multirow{2}{*}{RoI graph}            & \checkmark         &     &96     & 38.5 & 59.5 & 42.1 & 22.5 & 42   & 49.5 \\
       & \checkmark         &   \checkmark      &1.2k  & 38.9 & 60.3 & 42.2 & 23.2 & 42.4 & 49.8 \\
    \bottomrule
    \end{tabular}
\end{table}

\begin{table}[t!]
\begin{minipage}[t!]{.495\textwidth}\centering
    \caption{\textbf{Combination ablation}.  P: pixel relation, S: scale relation, R: RoI relation}
    \label{tab:ablation}    
    \resizebox{.98\linewidth}{!}{%
    \begin{tabular}{@{}lcccccc@{}}
    \toprule
                & AP   & AP$_{50}$ & AP$_{75}$ & AP$_\text{S}$  & AP$_\text{M}$  & AP$_\text{L}$  \\
    \midrule            
    P$+$S   &  38.7 & 59.5 & 41.9 & 23.3 & 42.0   & 50.1 \\
    S$+$R   & 39.4 & 60.7 & 42.7 & 23.3 & 42.4 & 51.8 \\
    P$+$R    & 39.5 & 60.5 & 43.1 & 23.7 & 42.9 & 51.1 \\
    P$+$S$+$R  & 39.4 & 60.3 & 43.1 & 23.4 & 42.6 & 51.5 \\
    \midrule                  
    Cascade RCNN & 40.4 & 60.2 & 43.8 & 23.8 & 44.1 & 52.2 \\
    SR + PR           & 41.5 & 61.9 & 45   & 25.5 & 45   & 54.1 \\
    PR + SR  &  \textbf{41.6} & \textbf{61.8} & \textbf{45.2} & \textbf{25.0}   & \textbf{45.1} & \textbf{54.2} \\
    \bottomrule
    \end{tabular}
    }
\end{minipage}
\hfill
\begin{minipage}[t!]{.495\textwidth}\centering
    \captionof{table}{HR-RCNN component ablation for \textbf{sharing box heads}~(denoted as 'Share') and \textbf{hierarchical reasoning}~(denoted as 'HR')}
    \label{tab:further-ablation1}
    \resizebox{.98\linewidth}{!}{%
    \begin{tabular}{@{}l|cc|cccccc@{}}
    \toprule
                & Share & HR  & AP   & AP$_{50}$ & AP$_{75}$ & AP$_\text{S}$  & AP$_\text{M}$  & AP$_\text{L}$  \\
    \midrule       
    HR-RCNN  &\checkmark &\checkmark       & \textbf{41.6} & \textbf{61.8} & \textbf{45.2} & \textbf{25}   & \textbf{45.1} & \textbf{54.2} \\
    w/o HR  &\checkmark &     & 40.2 & 60.1 & 43.5 & 23.5 & 43.7 & 52.8 \\
    w/o sharing head  & &\checkmark & 41.1 & 61.1 & 44.5 & 24.9 & 44.4 & 53.5 \\
    Cascade RCNN  & &  & 40.4 & 60.2 & 43.8 & 23.8 & 44.1 & 52.2 \\
    \bottomrule
    \end{tabular}
    }
\end{minipage}

\end{table}

\begin{table*}[t]
\caption{\textbf{Comparison with state-of-the-arts} on COCO test-dev. $^\dagger$: multi-scale testing, $^+$: with soft-NMS, large-batch BN}
\label{tab:sota}
\centering
\setlength{\cmidrulewidth}{0.01em}
\footnotesize
\renewcommand{\arraystretch}{1.2}
\renewcommand{\tabcolsep}{6pt}
\resizebox{.95\textwidth}{!}{
\begin{tabular}{@{}lllll cccccc@{}}
\toprule
Method   &Train size &Test size              & Backbone      & Epochs & AP   & AP$_{50}$ & AP$_{75}$ & AP$_\text{S}$  & AP$_\text{M}$  & AP$_\text{L}$  \\
\midrule
One-stage models:\\
\cmidrule[\cmidrulewidth]{1-11}
FCOS~\cite{tian2019fcos} & $1333\times800$ & $1333\times800$                & R101-FPN      & 24     & 41.5 & 60.7 & 45.0   & 24.4 & 44.8 & 51.6 \\
SAPD~\cite{zhu2019soft} & $1333\times800$ & $1333\times800$         & R101-FPN      & 24     & 43.5 & 63.6 & 46.5 & 24.9 & 46.8 & 54.6 \\
PAA~\cite{kim2020probabilistic} & $1333\times800$ & $1333\times800$       & R101-FPN      & 24     & 44.8 & 63.3 & 48.7 & 26.5 & 48.8 & 56.3 \\
MAL~\cite{ke2020multiple} & $1333\times800$ & $1333\times800$         & R101-FPN      & 24     & 43.6 & 62.8 & 47.1 & 25.0   & 46.9 & 55.8 \\
ATSS~\cite{zhang2020bridging} & $1333\times800$ & $1333\times800$             & R101-FPN      & 24     & 43.6 & 62.1 & 47.4 & 26.1 & 47.0   & 53.6 \\
SEPC~\cite{wang2020scale}& $1333\times800$ & $1333\times800$        & R101-FPN      & 24     & 45.5 & 64.9 & 49.5 & 27.0   & 48.8 & 56.7 \\
Detr~\cite{carion2020end} & $1333\times800$ & $1333\times800$        & R101-FPN      & 500    & 43.5 & 63.8 & 46.4 & 21.9 & 48.0   & 61.8 \\
BorderDet~\cite{qiu2020borderdet}  & $1333\times800$ & $1333\times800^\dagger$     & R101-DCN-FPN      & 24     & 47.2 & 66.1 & 51.0 & 28.1 & 50.2 & 59.9 \\ 
DDB-Net~\cite{chen2020dive} & $1333\times800$ & $1333\times800$        & R101-FPN      & 24     & 42.0   & 61.0   & 45.1 & 24.2 & 45.0   & 53.3 \\
DyHead~\cite{dai2021dyhead} & $1333\times800$ & $1333\times800$   &R101-FPN & 24 & 46.5 & 64.5 & 50.7 &28.3 &50.3 &57.3 \\
VFNet~\cite{zhang2021varifocalnet} & $1333\times800$ & $1333\times800$   &R101-FPN & 24 & 46.0 & 64.2 & 50.0 &27.5 & 49.4 & 56.9 \\
HoughNet\cite{HoughNet} & $512\times512$ & $512\times512^\dagger$     & HG-104      & 140     & 46.4   & 65.1   & 50.7 & 29.1 & 48.5   & 58.1 \\
RelationNet++\cite{relationnetplusplus2020} & $1333\times800$ &$1333\times800^\dagger$     & ResNeXt-$64 \times4$d-101-DCN     &  20     & \textbf{52.7}   & \textbf{70.4 }  & \textbf{58.3} & \textbf{35.8} & \textbf{55.3}  & \textbf{64.7} \\
\midrule
Two-stage models:\\
\cmidrule[\cmidrulewidth]{1-11}
DCN-V2~\cite{zhu2019deformable}  & $1333\times800$ & $1333\times800$                & R101-DCN-FPN  & 12     & 42.7 & 63.7 & 46.8 & 24.9 & 46.7 & 56.8 \\
Cascade RCNN~\cite{cai2018cascade}  & $1312\times800$ & $1312\times800$        & R101-FPN      & 18     & 42.8 & 62.1 & 46.3 & 23.7 & 45.5 & 55.2 \\
RPDet~\cite{yang2019reppoints} & $1333\times800$ & $1333\times800^\dagger$                  & R101-DCN-FPN & 12     & 46.5 & 67.4   & 50.9 & 30.3 & 49.7 & 57.1 \\ 
Cascade RCNN + SABL~\cite{wang2019side} & $1333\times800$ & $1333\times800$       & R101-FPN      & 12     & 43.3 & 60.9 & 46.2 & 23.8 & 46.5 & 55.7 \\
Dynamic RCNN~\cite{DynamicRCNN} & $1333\times800$ & $1333\times800^\dagger$     & R101-DCN-FPN      & 36     & 49.2   & 68.6 & 54.0 & 32.5 & 51.7 & 60.3 \\
AugFPN~\cite{fan2020augfpn}  & $1312\times800$ & $1312\times800$      & R101-augFPN   & 24     & 41.5 & 63.9 & 45.1 & 23.8 & 44.7 & 52.8 \\
TridenNet$^+$~\cite{li2019scale} & $1333\times800$ & $1333\times800$      & R101-DCN        & 36 &46.8 & 67.6 &51.5 & 28 &51.2 &60.5 \\
\cmidrule[\cmidrulewidth]{1-11}
HR-RCNN (Ours)         & $1333\times800$ & $1333\times800$     & R101-FPN      & 12     & 43.1 & 63.3 & 46.5 & 25.2 & 46.0   & 54.1 \\
HR-RCNN (Ours)        & $1333\times800$ & $1333\times800$      & R101-FPN      & 24     & 44.9 & 65.1 & 48.4 & 26.7 & 47.6 & 56.5 \\
HR-RCNN (Ours)        & $1333\times800$ & $1333\times800$    & R101-DCN-FPN  & 12     & 44.8 & 65.0   & 48.2 & 26.1 & 47.6 & 57.0   \\
HR-RCNN (Ours)      & $1333\times800$ & $1333\times800$      & R101-DCN-FPN  & 24     & 45.7 & 65.7 & 49.3 & 27.3 & 48.5 & 57.6 \\
HR-RCNN (Ours)     & $1333\times800$ & $1333\times800^\dagger$      & R101-DCN-FPN  & 24     & 47.7 & 68.2 & 51.7 & 30.8 & 50.4 & 59.4     \\
\bottomrule
\end{tabular}
}
\end{table*}

Firstly, we show the results of each single relational reasoning component in Tab.~\ref{tab:single-relation}. RoI relational reasoning brings the highest +0.9 improvement for AP, while pixel and scale reasoning improves by +0.5 each. Note that the extra parameters are quite marginal.

Then, we demonstrate the results of different combinations for different relation modules. As can be seen in  Tab.~\ref{tab:ablation}, combinations of any two reasoning components are complementary and consistently improve the performance, where +1.4/+1.5 gain come from scale-RoI reasoning and pixel-RoI reasoning respectively. But using all three (pixel-scale-RoI relation) jointly doesn't improve the performance any further. Finally, our HR-RCNN framework is able to successfully utilize these reasoning components hierarchically and performs the best.

Finally, we present an ablation study on the importance of weight sharing and hierarchical reasoning in Tab.~\ref{tab:further-ablation1}. Effectively, our approach without hierarchical reasoning is  a two-stage Cascade RCNN with shared box heads. Note that this drops the AP by 1.4 points. Next, similar to Cascade RCNN, we remove weight sharing and train disjoint the parameters for all branches. This not only increases the model parameters and inference time, but also hurts the AP by 0.5 points; further attesting to the importance of multitask hierarchical reasoning formulation.
Finally, compared with the original Cascade RCNN, HR-RCNN performs better by 1.2 points.
Note that though weight sharing helps in HR-RCNN, it hurts Cascade RCNN by 0.2 points. This supports our claim that hierarchical relationships can be fused implicitly through the shared box head.

\subsection{Comparison with State-of-the-art}

Tab.~\ref{tab:sota} shows comparisons of HR-RCNN with some state-of-the-art methods on the COCO test-dev split. 
Without bells and whistles, our HR-RCNN with ResNet-101 backbone achieves 44.9 AP using the 2x training scheme. By changing the backbone to deformable ResNet-101, HR-RCNN gets 47.7 mAP with a multi-scale testing. As a two-stage detector, HR-RCNN is comparable with all other two-stage methods under the same setup, \eg, with the same backbone and training epochs. Specifically, for pixel reasoning baseline DCN-V2~\cite{zhu2019deformable}, our hierarchical reasoning improves its AP by 2.1 points.
As a refinement baseline, Cascade RCNN refines the region proposals with three stages of box heads, while our HR-RCNN only have two stages of refinement and a single copy of box head weights. Although we use fewer refine steps, our HR-RCNN still outperforms Cascade RCNN with a considerable margin without adding more box heads. Note that our hierarchical visual reasoning is fairly generic and can be added to many backbones to further boost their performance.

\subsection{Experimental analysis}
\begin{figure}[t!]
    \centering
    \includegraphics[width=.95\linewidth]{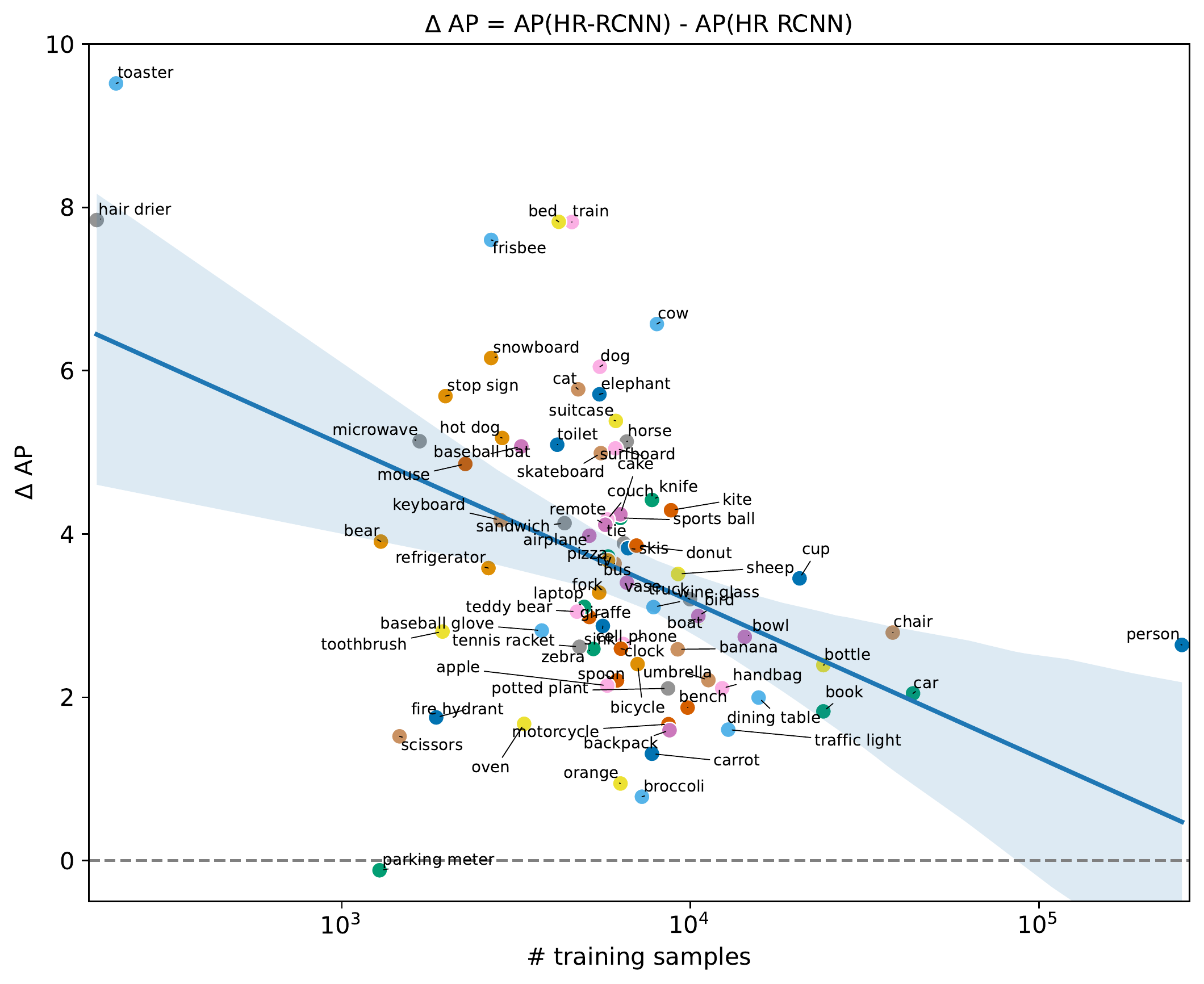}
    \caption{$\Delta$AP \vs training samples. Dots are AP gains by HR-CNN for different classes and we also show a linear regression fit. Nearly all classes can benefit from hierarchical reasoning, especially for in-frequent ones.}
    \label{fig:fig-class}
\end{figure}

To get a better understanding of HR-RCNN, we analysis its performance to number of training samples. We sort categories by the number of training samples, and evaluate how hierarchical reasoning behaves for frequent categories and rare categories. Compared to Faster R-CNN,
Fig~\ref{fig:fig-class} demonstrates the relationship between AP improvement and number of training samples, and we fit a linear regression line. HR-RCNN improves for all but one classes, with generally larger boost for in-frequent categories. In-frequent classes do not have enough training samples, and thus benefits more from hierarchical visual reasoning.

\section{Conclusion.}
In this paper, we propose a hierarchical relational reasoning framework (HR-RCNN) for object detection, which utilizes a concise graph attention module (GAM) to enable visual reasoning across heterogeneous nodes. We also explore different strategies to define the hierarchy between heterogeneous relationships, which leads to our HR-RCNN architecture. Finally, extensive experiments on COCO dataset show its effectiveness.

\paragraph{Acknowledgement.}
This work was supported by the Air Force STTR award (FA864920C0010), DARPA SemaFor program (HR001120C0124), DARPA SAIL-ON program (W911NF2020009), and Amazon Research Award to AS.

\bibliography{egbib}

\begin{thebibliography}{86}
\providecommand{\natexlab}[1]{#1}
\providecommand{\url}[1]{\texttt{#1}}
\expandafter\ifx\csname urlstyle\endcsname\relax
  \providecommand{\doi}[1]{doi: #1}\else
  \providecommand{\doi}{doi: \begingroup \urlstyle{rm}\Url}\fi

\bibitem[Arnab et~al.(2016)Arnab, Jayasumana, Zheng, and Torr]{arnab2016higher}
Anurag Arnab, Sadeep Jayasumana, Shuai Zheng, and Philip~HS Torr.
\newblock Higher order conditional random fields in deep neural networks.
\newblock In \emph{ECCV}, 2016.

\bibitem[Arnab et~al.(2018)Arnab, Zheng, Jayasumana, Romera-Paredes, Larsson,
  Kirillov, Savchynskyy, Rother, Kahl, and Torr]{arnab2018conditional}
Anurag Arnab, Shuai Zheng, Sadeep Jayasumana, Bernardino Romera-Paredes,
  M{\aa}ns Larsson, Alexander Kirillov, Bogdan Savchynskyy, Carsten Rother,
  Fredrik Kahl, and Philip~HS Torr.
\newblock Conditional random fields meet deep neural networks for semantic
  segmentation: Combining probabilistic graphical models with deep learning for
  structured prediction.
\newblock \emph{IEEE Signal Processing Magazine}, 2018.

\bibitem[Battaglia et~al.(2016)Battaglia, Pascanu, Lai, Rezende,
  et~al.]{battaglia2016interaction}
Peter Battaglia, Razvan Pascanu, Matthew Lai, Danilo~Jimenez Rezende, et~al.
\newblock Interaction networks for learning about objects, relations and
  physics.
\newblock In \emph{NeurIPS}, 2016.

\bibitem[Bell et~al.(2016)Bell, Lawrence~Zitnick, Bala, and
  Girshick]{bell2016inside}
Sean Bell, C~Lawrence~Zitnick, Kavita Bala, and Ross Girshick.
\newblock Inside-outside net: Detecting objects in context with skip pooling
  and recurrent neural networks.
\newblock In \emph{Proceedings of the IEEE conference on computer vision and
  pattern recognition}, pages 2874--2883, 2016.

\bibitem[Biederman(1981)]{biederman1981semantics}
Irving Biederman.
\newblock On the semantics of a glance at a scene.
\newblock \emph{Perceptual organization}, 213:\penalty0 253, 1981.

\bibitem[Cai and Vasconcelos(2018)]{cai2018cascade}
Zhaowei Cai and Nuno Vasconcelos.
\newblock Cascade r-cnn: Delving into high quality object detection.
\newblock In \emph{CVPR}, 2018.

\bibitem[Cao et~al.(2019)Cao, Xu, Lin, Wei, and Hu]{cao2019gcnet}
Yue Cao, Jiarui Xu, Stephen Lin, Fangyun Wei, and Han Hu.
\newblock Gcnet: Non-local networks meet squeeze-excitation networks and
  beyond.
\newblock \emph{arXiv}, 2019.

\bibitem[Carion et~al.(2020)Carion, Massa, Synnaeve, Usunier, Kirillov, and
  Zagoruyko]{carion2020end}
Nicolas Carion, Francisco Massa, Gabriel Synnaeve, Nicolas Usunier, Alexander
  Kirillov, and Sergey Zagoruyko.
\newblock End-to-end object detection with transformers.
\newblock 2020.

\bibitem[Chen et~al.(2020)Chen, Liu, Zhang, Liu, Yu, and Tai]{chen2020dive}
Ran Chen, Yong Liu, Mengdan Zhang, Shu Liu, Bei Yu, and Yu-Wing Tai.
\newblock Dive deeper into box for object detection.
\newblock 2020.

\bibitem[Chen and Gupta(2017)]{chen2017spatial}
Xinlei Chen and Abhinav Gupta.
\newblock Spatial memory for context reasoning in object detection.
\newblock In \emph{ICCV}, 2017.

\bibitem[Chen et~al.(2018)Chen, Li, Fei-Fei, and Gupta]{chen2018iterative}
Xinlei Chen, Li-Jia Li, Li~Fei-Fei, and Abhinav Gupta.
\newblock Iterative visual reasoning beyond convolutions.
\newblock In \emph{CVPR}, 2018.

\bibitem[Chi et~al.(2020)Chi, Wei, and Hu]{relationnetplusplus2020}
Cheng Chi, Fangyun Wei, and Han Hu.
\newblock Relationnet++: Bridging visual representations for object detection
  via transformer decoder.
\newblock In \emph{NeurIPS}, 2020.

\bibitem[Dai et~al.(2017)Dai, Qi, Xiong, Li, Zhang, Hu, and
  Wei]{dai2017deformable}
Jifeng Dai, Haozhi Qi, Yuwen Xiong, Yi~Li, Guodong Zhang, Han Hu, and Yichen
  Wei.
\newblock Deformable convolutional networks.
\newblock In \emph{ICCV}, 2017.

\bibitem[Dai et~al.(2021)Dai, Chen, Xiao, Chen, Liu, Yuan, and
  Zhang]{dai2021dyhead}
Xiyang Dai, Yinpeng Chen, Bin Xiao, Dongdong Chen, Mengchen Liu, Lu~Yuan, and
  Lei Zhang.
\newblock Dynamic head: Unifying object detection heads with attentions.
\newblock In \emph{Proceedings of the IEEE/CVF Conference on Computer Vision
  and Pattern Recognition (CVPR)}, pages 7373--7382, June 2021.

\bibitem[Fan et~al.(2020)Fan, Zhang, Xiang, Pan, et~al.]{fan2020augfpn}
Bin Fan, Qian Zhang, Shiming Xiang, Chunhong Pan, et~al.
\newblock Augfpn: Improving multi-scale feature learning for object detection.
\newblock 2020.

\bibitem[Fu et~al.(2017)Fu, Liu, Ranga, Tyagi, and Berg]{fu2017dssd}
Cheng-Yang Fu, Wei Liu, Ananth Ranga, Ambrish Tyagi, and Alexander~C Berg.
\newblock Dssd: Deconvolutional single shot detector.
\newblock \emph{arXiv preprint arXiv:1701.06659}, 2017.

\bibitem[Ghiasi et~al.(2019)Ghiasi, Lin, and Le]{ghiasi2019fpn}
Golnaz Ghiasi, Tsung-Yi Lin, and Quoc~V Le.
\newblock Nas-fpn: Learning scalable feature pyramid architecture for object
  detection.
\newblock In \emph{CVPR}, 2019.

\bibitem[Gu et~al.(2019)Gu, Zhao, Lin, Li, Cai, and Ling]{gu2019scene}
Jiuxiang Gu, Handong Zhao, Zhe Lin, Sheng Li, Jianfei Cai, and Mingyang Ling.
\newblock Scene graph generation with external knowledge and image
  reconstruction.
\newblock In \emph{CVPR}, 2019.

\bibitem[Hamaguchi et~al.(2017)Hamaguchi, Oiwa, Shimbo, and
  Matsumoto]{hamaguchi2017knowledge}
Takuo Hamaguchi, Hidekazu Oiwa, Masashi Shimbo, and Yuji Matsumoto.
\newblock Knowledge transfer for out-of-knowledge-base entities: A graph neural
  network approach.
\newblock 2017.

\bibitem[Hamilton et~al.(2017)Hamilton, Ying, and
  Leskovec]{hamilton2017inductive}
Will Hamilton, Zhitao Ying, and Jure Leskovec.
\newblock Inductive representation learning on large graphs.
\newblock In \emph{NeurIPS}, 2017.

\bibitem[Hariharan et~al.(2015)Hariharan, Arbel{\'a}ez, Girshick, and
  Malik]{hariharan2015hypercolumns}
Bharath Hariharan, Pablo Arbel{\'a}ez, Ross Girshick, and Jitendra Malik.
\newblock Hypercolumns for object segmentation and fine-grained localization.
\newblock In \emph{CVPR}, 2015.

\bibitem[{He} et~al.(2016){He}, {Zhang}, {Ren}, and {Sun}]{resnet}
Kaiming {He}, Xiangyu {Zhang}, Shaoqing {Ren}, and Jian {Sun}.
\newblock Deep residual learning for image recognition.
\newblock In \emph{CVPR}, 2016.

\bibitem[He et~al.(2017)He, Gkioxari, Doll{\'a}r, and Girshick]{he2017mask}
Kaiming He, Georgia Gkioxari, Piotr Doll{\'a}r, and Ross Girshick.
\newblock Mask r-cnn.
\newblock In \emph{ICCV}, 2017.

\bibitem[Hock et~al.(1974)Hock, Gordon, and Whitehurst]{hock1974contextual}
Howard~S Hock, Gregory~P Gordon, and Robert Whitehurst.
\newblock Contextual relations: The influence of familiarity, physical
  plausibility, and belongingness.
\newblock \emph{Perception \& Psychophysics}, 16\penalty0 (1):\penalty0 4--8,
  1974.

\bibitem[Hu et~al.(2018)Hu, Gu, Zhang, Dai, and Wei]{hu2018relation}
Han Hu, Jiayuan Gu, Zheng Zhang, Jifeng Dai, and Yichen Wei.
\newblock Relation networks for object detection.
\newblock In \emph{CVPR}, 2018.

\bibitem[Huang et~al.(2019)Huang, Wang, Huang, Huang, Wei, and
  Liu]{huang2018ccnet}
Zilong Huang, Xinggang Wang, Lichao Huang, Chang Huang, Yunchao Wei, and Wenyu
  Liu.
\newblock Ccnet: Criss-cross attention for semantic segmentation.
\newblock In \emph{ICCV}, 2019.

\bibitem[Jiang et~al.(2018)Jiang, Xu, Liang, and Lin]{jiang2018hybrid}
Chenhan Jiang, Hang Xu, Xiaodan Liang, and Liang Lin.
\newblock Hybrid knowledge routed modules for large-scale object detection.
\newblock In \emph{NeurIPS}, 2018.

\bibitem[Ke et~al.(2020)Ke, Zhang, Huang, Ye, Liu, and Huang]{ke2020multiple}
Wei Ke, Tianliang Zhang, Zeyi Huang, Qixiang Ye, Jianzhuang Liu, and Dong
  Huang.
\newblock Multiple anchor learning for visual object detection.
\newblock In \emph{CVPR}, 2020.

\bibitem[Khalil et~al.(2017)Khalil, Dai, Zhang, Dilkina, and
  Song]{khalil2017learning}
Elias Khalil, Hanjun Dai, Yuyu Zhang, Bistra Dilkina, and Le~Song.
\newblock Learning combinatorial optimization algorithms over graphs.
\newblock In \emph{NeurIPS}, 2017.

\bibitem[Kim and Lee(2020)]{kim2020probabilistic}
Kang Kim and Hee~Seok Lee.
\newblock Probabilistic anchor assignment with iou prediction for object
  detection.
\newblock 2020.

\bibitem[Kipf and Welling(2017)]{kipf2016semi}
Thomas~N Kipf and Max Welling.
\newblock Semi-supervised classification with graph convolutional networks.
\newblock In \emph{ICLR}, 2017.

\bibitem[Kong et~al.(2017)Kong, Sun, Yao, Liu, Lu, and Chen]{kong2017ron}
Tao Kong, Fuchun Sun, Anbang Yao, Huaping Liu, Ming Lu, and Yurong Chen.
\newblock Ron: Reverse connection with objectness prior networks for object
  detection.
\newblock In \emph{CVPR}, 2017.

\bibitem[Kosaraju et~al.(2019)Kosaraju, Sadeghian, Mart{\'\i}n-Mart{\'\i}n,
  Reid, Rezatofighi, and Savarese]{kosaraju2019social}
Vineet Kosaraju, Amir Sadeghian, Roberto Mart{\'\i}n-Mart{\'\i}n, Ian Reid,
  Hamid Rezatofighi, and Silvio Savarese.
\newblock Social-bigat: Multimodal trajectory forecasting using bicycle-gan and
  graph attention networks.
\newblock In \emph{NeurIPS}, 2019.

\bibitem[{Krizhevsky} et~al.(2012){Krizhevsky}, {Sutskever}, and
  {Hinton}]{alexnet}
Alex {Krizhevsky}, Ilya {Sutskever}, and Geoffrey~E. {Hinton}.
\newblock {ImageNet} classification with deep convolutional neural networks.
\newblock In \emph{NeurIPS}, 2012.

\bibitem[{Krähenbühl} and {Koltun}(2011)]{krahenbuhl2011efficient}
Philipp {Krähenbühl} and Vladlen {Koltun}.
\newblock Efficient inference in fully connected crfs with gaussian edge
  potentials.
\newblock \emph{NeurIPS}, 2011.

\bibitem[Li et~al.(2019{\natexlab{a}})Li, Gan, Cheng, and Liu]{li2019relation}
Linjie Li, Zhe Gan, Yu~Cheng, and Jingjing Liu.
\newblock Relation-aware graph attention network for visual question answering.
\newblock In \emph{ICCV}, 2019{\natexlab{a}}.

\bibitem[Li et~al.(2019{\natexlab{b}})Li, Chen, Wang, and Zhang]{li2019scale}
Yanghao Li, Yuntao Chen, Naiyan Wang, and Zhaoxiang Zhang.
\newblock Scale-aware trident networks for object detection.
\newblock In \emph{ICCV}, 2019{\natexlab{b}}.

\bibitem[Li et~al.(2018)Li, Ouyang, Zhou, Shi, Zhang, and
  Wang]{li2018factorizable}
Yikang Li, Wanli Ouyang, Bolei Zhou, Jianping Shi, Chao Zhang, and Xiaogang
  Wang.
\newblock Factorizable net: an efficient subgraph-based framework for scene
  graph generation.
\newblock In \emph{ECCV)}, pages 335--351, 2018.

\bibitem[Li and Zhou(2017)]{li2017fssd}
Zuoxin Li and Fuqiang Zhou.
\newblock Fssd: feature fusion single shot multibox detector.
\newblock \emph{arXiv preprint arXiv:1712.00960}, 2017.

\bibitem[Lin et~al.(2014)Lin, Maire, Belongie, Hays, Perona, Ramanan,
  Doll{\'a}r, and Zitnick]{lin2014microsoft}
Tsung-Yi Lin, Michael Maire, Serge Belongie, James Hays, Pietro Perona, Deva
  Ramanan, Piotr Doll{\'a}r, and C~Lawrence Zitnick.
\newblock Microsoft coco: Common objects in context.
\newblock In \emph{ECCV}. Springer, 2014.

\bibitem[Lin et~al.(2017{\natexlab{a}})Lin, Dollar, Girshick, He, Hariharan,
  and Belongie]{Lin_2017_CVPR}
Tsung-Yi Lin, Piotr Dollar, Ross Girshick, Kaiming He, Bharath Hariharan, and
  Serge Belongie.
\newblock Feature pyramid networks for object detection.
\newblock In \emph{CVPR}, 2017{\natexlab{a}}.

\bibitem[Lin et~al.(2017{\natexlab{b}})Lin, Doll{\'a}r, Girshick, He,
  Hariharan, and Belongie]{lin2017feature}
Tsung-Yi Lin, Piotr Doll{\'a}r, Ross Girshick, Kaiming He, Bharath Hariharan,
  and Serge Belongie.
\newblock Feature pyramid networks for object detection.
\newblock In \emph{CVPR}, 2017{\natexlab{b}}.

\bibitem[Lin et~al.(2017{\natexlab{c}})Lin, Goyal, Girshick, He, and
  Doll{\'a}r]{lin2017focal}
Tsung-Yi Lin, Priya Goyal, Ross Girshick, Kaiming He, and Piotr Doll{\'a}r.
\newblock Focal loss for dense object detection.
\newblock In \emph{ICCV}, 2017{\natexlab{c}}.

\bibitem[Liu et~al.(2018{\natexlab{a}})Liu, Qi, Qin, Shi, and Jia]{liu2018path}
Shu Liu, Lu~Qi, Haifang Qin, Jianping Shi, and Jiaya Jia.
\newblock Path aggregation network for instance segmentation.
\newblock In \emph{CVPR}, 2018{\natexlab{a}}.

\bibitem[Liu et~al.(2018{\natexlab{b}})Liu, Wang, Shan, and
  Chen]{liu2018structure}
Yong Liu, Ruiping Wang, Shiguang Shan, and Xilin Chen.
\newblock Structure inference net: Object detection using scene-level context
  and instance-level relationships.
\newblock In \emph{CVPR}, 2018{\natexlab{b}}.

\bibitem[Marino et~al.(2016)Marino, Salakhutdinov, and Gupta]{marino2016more}
Kenneth Marino, Ruslan Salakhutdinov, and Abhinav Gupta.
\newblock The more you know: Using knowledge graphs for image classification.
\newblock \emph{arXiv preprint arXiv:1612.04844}, 2016.

\bibitem[Newell et~al.(2016)Newell, Yang, and Deng]{newell2016stacked}
Alejandro Newell, Kaiyu Yang, and Jia Deng.
\newblock Stacked hourglass networks for human pose estimation.
\newblock In \emph{ECCV}, 2016.

\bibitem[Palmer(1975)]{palmer1975effects}
tephen~E Palmer.
\newblock The effects of contextual scenes on the identification of objects.
\newblock \emph{Memory \& Cognition}, 3:\penalty0 519--526, 1975.

\bibitem[Qiao et~al.(2020)Qiao, Chen, and Yuille]{qiao2020detectors}
Siyuan Qiao, Liang-Chieh Chen, and Alan Yuille.
\newblock Detectors: Detecting objects with recursive feature pyramid and
  switchable atrous convolution.
\newblock \emph{arXiv preprint arXiv:2006.02334}, 2020.

\bibitem[Qiu et~al.(2020)Qiu, Ma, Li, Liu, and Sun]{qiu2020borderdet}
Han Qiu, Yuchen Ma, Zeming Li, Songtao Liu, and Jian Sun.
\newblock Borderdet: Border feature for dense object detection.
\newblock 2020.

\bibitem[Ren et~al.(2015)Ren, He, Girshick, and Sun]{ren2015faster}
Shaoqing Ren, Kaiming He, Ross Girshick, and Jian Sun.
\newblock Faster r-cnn: Towards real-time object detection with region proposal
  networks.
\newblock In \emph{NeurIPS}, 2015.

\bibitem[Samet et~al.(2020)Samet, Hicsonmez, and Akbas]{HoughNet}
Nermin Samet, Samet Hicsonmez, and Emre Akbas.
\newblock Houghnet: Integrating near and long-range evidence for bottom-up
  object detection.
\newblock In \emph{European Conference on Computer Vision (ECCV)}, 2020.

\bibitem[Sanchez-Gonzalez et~al.(2018)Sanchez-Gonzalez, Heess, Springenberg,
  Merel, Riedmiller, Hadsell, and Battaglia]{sanchez2018graph}
Alvaro Sanchez-Gonzalez, Nicolas Heess, Jost~Tobias Springenberg, Josh Merel,
  Martin Riedmiller, Raia Hadsell, and Peter Battaglia.
\newblock Graph networks as learnable physics engines for inference and
  control.
\newblock \emph{arXiv preprint arXiv:1806.01242}, 2018.

\bibitem[Shrivastava and Gupta(2016)]{primingfeedback_eccv16}
Abhinav Shrivastava and Abhinav Gupta.
\newblock Contextual {P}riming and {F}eedback for {F}aster {R-CNN}.
\newblock In \emph{European Conference on Computer Vision (ECCV)}, 2016.

\bibitem[Shrivastava et~al.(2016{\natexlab{a}})Shrivastava, Gupta, and
  Girshick]{ohem_cvpr16}
Abhinav Shrivastava, Abhinav Gupta, and Ross Girshick.
\newblock {Training Region-based Object Detectors with Online Hard Example
  Mining}.
\newblock In \emph{{IEEE Conference on Computer Vision and Pattern Recognition
  (CVPR)}}, 2016{\natexlab{a}}.

\bibitem[Shrivastava et~al.(2016{\natexlab{b}})Shrivastava, Sukthankar, Malik,
  and Gupta]{shrivastava2016beyond}
Abhinav Shrivastava, Rahul Sukthankar, Jitendra Malik, and Abhinav Gupta.
\newblock Beyond skip connections: Top-down modulation for object detection.
\newblock \emph{arXiv preprint arXiv:1612.06851}, 2016{\natexlab{b}}.

\bibitem[{Simonyan} and {Zisserman}(2015)]{vgg}
Karen {Simonyan} and Andrew {Zisserman}.
\newblock Very deep convolutional networks for large-scale image recognition.
\newblock In \emph{ICLR}, 2015.

\bibitem[Singh and Davis(2018)]{singh2018analysis}
Bharat Singh and Larry~S Davis.
\newblock An analysis of scale invariance in object detection snip.
\newblock In \emph{CVPR}, 2018.

\bibitem[Singh et~al.(2018)Singh, Najibi, and Davis]{singh2018sniper}
Bharat Singh, Mahyar Najibi, and Larry~S Davis.
\newblock Sniper: Efficient multi-scale training.
\newblock In \emph{NeurIPS}, 2018.

\bibitem[Sun et~al.(2019{\natexlab{a}})Sun, Xiao, Liu, and Wang]{sun2019deep}
Ke~Sun, Bin Xiao, Dong Liu, and Jingdong Wang.
\newblock Deep high-resolution representation learning for human pose
  estimation.
\newblock In \emph{CVPR}, 2019{\natexlab{a}}.

\bibitem[Sun et~al.(2019{\natexlab{b}})Sun, Zhao, Jiang, Cheng, Xiao, Liu, Mu,
  Wang, Liu, and Wang]{sun2019high}
Ke~Sun, Yang Zhao, Borui Jiang, Tianheng Cheng, Bin Xiao, Dong Liu, Yadong Mu,
  Xinggang Wang, Wenyu Liu, and Jingdong Wang.
\newblock High-resolution representations for labeling pixels and regions.
\newblock \emph{arXiv preprint arXiv:1904.04514}, 2019{\natexlab{b}}.

\bibitem[Tan et~al.(2020)Tan, Pang, and Le]{Tan_2020_CVPR}
Mingxing Tan, Ruoming Pang, and Quoc~V. Le.
\newblock Efficientdet: Scalable and efficient object detection.
\newblock In \emph{CVPR}, 2020.

\bibitem[Tian et~al.(2019)Tian, Shen, Chen, and He]{tian2019fcos}
Zhi Tian, Chunhua Shen, Hao Chen, and Tong He.
\newblock Fcos: Fully convolutional one-stage object detection.
\newblock In \emph{ICCV}, 2019.

\bibitem[Torralba et~al.(2003)Torralba, Murphy, Freeman, Rubin,
  et~al.]{torralba2003context}
Antonio Torralba, Kevin~P Murphy, William~T Freeman, Mark~A Rubin, et~al.
\newblock Context-based vision system for place and object recognition.
\newblock In \emph{ICCV}, volume~3, pages 273--280, 2003.

\bibitem[Veli{\v{c}}kovi{\'c} et~al.(2017)Veli{\v{c}}kovi{\'c}, Cucurull,
  Casanova, Romero, Lio, and Bengio]{velivckovic2017graph}
Petar Veli{\v{c}}kovi{\'c}, Guillem Cucurull, Arantxa Casanova, Adriana Romero,
  Pietro Lio, and Yoshua Bengio.
\newblock Graph attention networks.
\newblock \emph{arXiv preprint arXiv:1710.10903}, 2017.

\bibitem[Wang et~al.(2020{\natexlab{a}})Wang, Zhang, Cao, Chen, Pang, Gong,
  Shi, Loy, and Lin]{wang2019side}
Jiaqi Wang, Wenwei Zhang, Yuhang Cao, Kai Chen, Jiangmiao Pang, Tao Gong,
  Jianping Shi, Chen~Change Loy, and Dahua Lin.
\newblock Side-aware boundary localization for more precise object detection.
\newblock 2020{\natexlab{a}}.

\bibitem[Wang et~al.(2019)Wang, Huang, Hou, Zhang, and Shan]{wang2019graph}
Lei Wang, Yuchun Huang, Yaolin Hou, Shenman Zhang, and Jie Shan.
\newblock Graph attention convolution for point cloud semantic segmentation.
\newblock In \emph{CVPR}, 2019.

\bibitem[Wang et~al.(2018)Wang, Girshick, Gupta, and He]{wang2018nonlocal}
Xiaolong Wang, Ross Girshick, Abhinav Gupta, and Kaiming He.
\newblock Non-local neural networks.
\newblock In \emph{CVPR}, 2018.

\bibitem[Wang et~al.(2020{\natexlab{b}})Wang, Zhang, Yu, Feng, and
  Zhang]{wang2020scale}
Xinjiang Wang, Shilong Zhang, Zhuoran Yu, Litong Feng, and Wayne Zhang.
\newblock Scale-equalizing pyramid convolution for object detection.
\newblock In \emph{CVPR}, 2020{\natexlab{b}}.

\bibitem[Woo et~al.(2019)Woo, Hwang, Jang, and Kweon]{woo2019gated}
Sanghyun Woo, Soonmin Hwang, Ho-Deok Jang, and In~So Kweon.
\newblock Gated bidirectional feature pyramid network for accurate one-shot
  detection.
\newblock \emph{Machine Vision and Applications}, 2019.

\bibitem[Wu et~al.(2019)Wu, Kirillov, Massa, Lo, and
  Girshick]{wu2019detectron2}
Yuxin Wu, Alexander Kirillov, Francisco Massa, Wan-Yen Lo, and Ross Girshick.
\newblock Detectron2.
\newblock \url{https://github.com/facebookresearch/detectron2}, 2019.

\bibitem[Xu et~al.(2017)Xu, Ouyang, Alameda-Pineda, Ricci, Wang, and
  Sebe]{xu2017learning}
Dan Xu, Wanli Ouyang, Xavier Alameda-Pineda, Elisa Ricci, Xiaogang Wang, and
  Nicu Sebe.
\newblock Learning deep structured multi-scale features using attention-gated
  crfs for contour prediction.
\newblock In \emph{NeurIPS}, 2017.

\bibitem[Xu et~al.(2019{\natexlab{a}})Xu, Jiang, Liang, and Li]{xu2019spatial}
Hang Xu, Chenhan Jiang, Xiaodan Liang, and Zhenguo Li.
\newblock Spatial-aware graph relation network for large-scale object
  detection.
\newblock In \emph{CVPR}, 2019{\natexlab{a}}.

\bibitem[Xu et~al.(2019{\natexlab{b}})Xu, Jiang, Liang, Lin, and
  Li]{xu2019reasoning}
Hang Xu, ChenHan Jiang, Xiaodan Liang, Liang Lin, and Zhenguo Li.
\newblock Reasoning-rcnn: Unifying adaptive global reasoning into large-scale
  object detection.
\newblock In \emph{CVPR}, 2019{\natexlab{b}}.

\bibitem[Yang et~al.(2019{\natexlab{a}})Yang, Li, and Yu]{yang2019dynamic}
Sibei Yang, Guanbin Li, and Yizhou Yu.
\newblock Dynamic graph attention for referring expression comprehension.
\newblock In \emph{ICCV}, 2019{\natexlab{a}}.

\bibitem[Yang et~al.(2019{\natexlab{b}})Yang, Liu, Hu, Wang, and
  Lin]{yang2019reppoints}
Ze~Yang, Shaohui Liu, Han Hu, Liwei Wang, and Stephen Lin.
\newblock Reppoints: Point set representation for object detection.
\newblock In \emph{ICCV}, 2019{\natexlab{b}}.

\bibitem[Yu et~al.(2018)Yu, Wang, Shelhamer, and Darrell]{yu2018deep}
Fisher Yu, Dequan Wang, Evan Shelhamer, and Trevor Darrell.
\newblock Deep layer aggregation.
\newblock In \emph{CVPR}, 2018.

\bibitem[Zellers et~al.(2018)Zellers, Yatskar, Thomson, and
  Choi]{zellers2018neural}
Rowan Zellers, Mark Yatskar, Sam Thomson, and Yejin Choi.
\newblock Neural motifs: Scene graph parsing with global context.
\newblock In \emph{CVPR}, 2018.

\bibitem[Zhang et~al.(2021)Zhang, Wang, Dayoub, and
  S{\"u}nderhauf]{zhang2021varifocalnet}
Haoyang Zhang, Ying Wang, Feras Dayoub, and Niko S{\"u}nderhauf.
\newblock Varifocalnet: An iou-aware dense object detector.
\newblock In \emph{CVPR}, 2021.

\bibitem[Zhang et~al.(2020{\natexlab{a}})Zhang, Chang, Ma, Wang, and
  Chen]{DynamicRCNN}
Hongkai Zhang, Hong Chang, Bingpeng Ma, Naiyan Wang, and Xilin Chen.
\newblock Dynamic {R-CNN}: Towards high quality object detection via dynamic
  training.
\newblock 2020{\natexlab{a}}.

\bibitem[Zhang et~al.(2020{\natexlab{b}})Zhang, Xu, Arnab, and
  Torr]{Zhang2020DynamicGM}
Li~Zhang, D.~Xu, A.~Arnab, and P.~Torr.
\newblock Dynamic graph message passing networks.
\newblock 2020{\natexlab{b}}.

\bibitem[Zhang et~al.(2020{\natexlab{c}})Zhang, Chi, Yao, Lei, and
  Li]{zhang2020bridging}
Shifeng Zhang, Cheng Chi, Yongqiang Yao, Zhen Lei, and Stan~Z Li.
\newblock Bridging the gap between anchor-based and anchor-free detection via
  adaptive training sample selection.
\newblock In \emph{CVPR}, 2020{\natexlab{c}}.

\bibitem[Zhao et~al.(2019)Zhao, Sheng, Wang, Tang, Chen, Cai, and
  Ling]{zhao2019m2det}
Qijie Zhao, Tao Sheng, Yongtao Wang, Zhi Tang, Ying Chen, Ling Cai, and Haibin
  Ling.
\newblock M2det: A single-shot object detector based on multi-level feature
  pyramid network.
\newblock In \emph{AAAI}, 2019.

\bibitem[{Zheng} et~al.(2015){Zheng}, {Jayasumana}, {Romera-Paredes}, {Vineet},
  {Su}, {Du}, {Huang}, and {Torr}]{zheng2015conditional}
Shuai {Zheng}, Sadeep {Jayasumana}, Bernardino {Romera-Paredes}, Vibhav
  {Vineet}, Zhizhong {Su}, Dalong {Du}, Chang {Huang}, and Philip H.~S. {Torr}.
\newblock Conditional random fields as recurrent neural networks.
\newblock In \emph{ICCV}, 2015.

\bibitem[Zhu et~al.(2020)Zhu, Chen, Shen, and Savvides]{zhu2019soft}
Chenchen Zhu, Fangyi Chen, Zhiqiang Shen, and Marios Savvides.
\newblock Soft anchor-point object detection.
\newblock 2020.

\bibitem[Zhu et~al.(2019)Zhu, Hu, Lin, and Dai]{zhu2019deformable}
Xizhou Zhu, Han Hu, Stephen Lin, and Jifeng Dai.
\newblock Deformable convnets v2: More deformable, better results.
\newblock In \emph{CVPR}, 2019.

\end{thebibliography}

\end{document}